\documentclass[letterpaper]{article} 
\usepackage{aaai24}  
\usepackage{times}  
\usepackage{helvet}  
\usepackage{courier}  
\usepackage[hyphens]{url}  
\usepackage{graphicx} 
\usepackage{natbib}  
\usepackage{caption} 
\usepackage{algorithm}
\usepackage{algorithmic}

\usepackage{multirow}
\usepackage{rotating}
\usepackage{amsmath}
\usepackage{amssymb}

\usepackage{newfloat}
\usepackage{listings}
\DeclareCaptionStyle{ruled}{labelfont=normalfont,labelsep=colon,strut=off} 
\floatstyle{ruled}
\newfloat{listing}{tb}{lst}{}
\floatname{listing}{Listing}
\title{Relightable and Animatable Neural Avatars from Videos}
\author {
    Wenbin Lin,
    Chengwei Zheng,
    Jun-Hai Yong, 
    Feng Xu
}
\affiliations {
    School of Software and BNRist, Tsinghua University \\
    lwb20@mails.tsinghua.edu.cn, zhengcw18@gmail.com, yongjh@tsinghua.edu.cn, xufeng2003@gmail.com
}

\usepackage{bibentry}

\begin{document}

\maketitle

\begin{abstract}

Lightweight creation of 3D digital avatars is a highly desirable but challenging task.
With only sparse videos of a person under unknown illumination, we propose a method to create relightable and animatable neural avatars, which can be used to synthesize photorealistic images of humans under novel viewpoints, body poses, and lighting.
The key challenge here is to disentangle the geometry, material of the clothed body, and lighting, which becomes more difficult due to the complex geometry and shadow changes caused by body motions. 
To solve this ill-posed problem, we propose novel techniques to better model the geometry and shadow changes. 
For geometry change modeling, we propose an invertible deformation field, which helps to solve the inverse skinning problem and leads to better geometry quality.
To model the spatial and temporal varying shading cues, we propose a pose-aware part-wise light visibility network to estimate light occlusion.
Extensive experiments on synthetic and real datasets show that our approach reconstructs high-quality geometry and generates realistic shadows under different body poses.
Code and data are available at \url{https://wenbin-lin.github.io/RelightableAvatar-page/}.

\end{abstract}
\section{Introduction}
Human digitizing has been rapidly developed in recent years, in which the reconstruction and animation of 3D clothed human avatars have many applications in telepresence, AR/VR, and virtual try-on.
One important goal here is to render the human avatar in desired lighting environment with desired poses.
Therefore, the human avatars need to be both relightable and animatable and achieve photorealistic rendering quality.
Usually, the generation of these high-quality human avatars relies on high-quality data like the ones recorded by Light Stages~\cite{lightstage} which are complicated and expensive. 

Recently, the emergence of Neural Radiance Fileds (NeRF) \cite{mildenhall2020nerf} opens a new window to generate animatable and relightable 3D human avatars just from the daily recorded videos.
NeRF-based methods have achieved remarkable success in 3D object representation and photorealistic rendering of both static and dynamic objects including human bodies \cite{neuralbody, animatablenerf, hnerf, humannerf, selfrecon, neuman, ARAH, peng2022animatable, monohuman, su2023npc}.
Also, NeRF can be used for intrinsic decomposition to achieve impressive relighting results for static objects \cite{NeRFactor, NeILF, NeRD, NeRV, Neural-PIL, InvRender, tensoir}.
However, NeRF-based dynamic object relighting is rarely studied. 
One key challenge is that the dynamics cause dramatic changes in object shading, which is hard to model with the current NeRF techniques. 

In this work, we propose to reconstruct both relightable and animatable 3D human avatars from sparse videos recorded under uncalibrated illuminations. 
To achieve this goal, we need to reconstruct the body geometry, material, and environmental light.
The dynamic body geometry is modeled by a static geometry in a canonical space and the motion to deform it to the shape in the observation space of each frame.
We propose an invertible neural deformation field that builds a bidirectional mapping between points of the canonical space and all observation spaces. 
With this bidirectional mapping, we can easily leverage the body mesh extracted in the canonical pose to better solve the inverse linear blend skinning problem, thus achieving high-quality geometry reconstruction.
After the geometry reconstruction of all frames, we propose a light visibility estimation module to better model the dynamic self-occlusion effect for material and light reconstruction.
We transfer the global pose-related visibility estimation task into multiple, part-wise, local ones, which dramatically simplifies the complexity of light visibility estimation.
This model has good generalization capability with limited training data benefiting from the part-wise architecture, and thus successfully estimates the light visibility under various body poses and lighting conditions.
Finally, we optimize the body material and lighting parameters, and then our method can render photorealistic images under any desired body pose, lighting, and viewpoint.
In summary, the contributions include:
\begin{itemize}
    \item 
    the first method that is able to reconstruct both relightable and animatable human avatars with plausible shadow effects from sparse multiview videos,
    \item 
    an invertible deformation field that better solves the inverse skinning problem, leading to accurate dense correspondence between different body poses,
    \item 
    part-wise light visibility networks that better estimate pose and light-related shading cues with high generalization capability.
\end{itemize}

\section{Related Work}

\subsection{Neural Human Avatars}
In recent years, neural radiance fields (NeRF) \cite{mildenhall2020nerf} have shown great abilities in photorealistic rendering. 
And many methods have successfully combined NeRF with human parametric models for human body reconstruction \cite{neuralbody, humannerf} and animatable human body modeling \cite{ARAH, neuman, slrf, ani-nerf, animatablenerf, peng2022animatable, selfrecon, monohuman} with sparse videos. 
For dynamic body motion modeling, people usually leverage linear blend skinning (LBS) \cite{pose_deformation} to drive the body to different poses and use neural displacement fields to model the non-rigid deformations.
Among these works, the deformation fields only model single-direction displacement, either forward deformation (canonical to observation) \cite{ARAH, tava} or backward deformation (observation to canonical) \cite{animatablenerf, ani-nerf, peng2022animatable}.
Different from them, our method proposes an invertible deformation field to solve the correspondence between canonical and observation space bidirectionally, which helps to better solve the inverse skinning problem, and leads to better geometry reconstruction.
Recent work MonoHuman \cite{monohuman} also models bidirectional deformations, but unlike the compact single invertible network in our approach, they use two non-invertible neural networks to model the deformations separately.
Additionally, these methods model body appearance using view-dependent color without decomposing it into lighting and reflectance.
In contrast, our method enables relighting by reconstructing the environment lighting and the surface material.

\subsection{Human Relighting}

Some methods have been proposed to enable relighting of human images \cite{sun2019single, wang2020single, zhou2019deep, kanamori2019relighting, pandey2021total, ji2022geometry}. However, these image-based methods do not support changing the viewpoints and human poses. 
To further enable novel view relighting, 3D reconstruction techniques have been leveraged to model the human geometry \cite{guo2019relightables}. 
For video-based human relighting, Relighting4D \cite{Relighting4D} enables free-viewpoint relighting from only human videos under unknown illuminations by using a set of neural fields of normal, occlusion, diffuse, and specular maps. 
But it is hard to relight the human with novel poses as it involves per-frame latent features which are not generalizable for novel poses. 
RANA \cite{rana} proposed a generalizable relightable articulated neural avatars creation method based on SMPL+D \cite{alldieck2018video} model with albedo, normal map refinement techniques. 
But their method did not model specular reflection and cast shadows.
In this paper, we present the first method that can reconstruct relightable and animatable human avatars from videos under unknown illuminations, while providing physically correct shadows.

\subsection{Invertible Neural Network}

Invertible Neural Networks (INNs) \cite{nice, real-nvp, i-resnet, neural-ode, glow} are are capable of performing invertible transformations between the input and output space.
They are widely used in generative models like Normalizing Flows \cite{normalizingflow} for density estimation.
Moreover, the ability of INNs to maintain cycle consistency between two spaces makes them suitable for modeling the deformation field of 3D objects. 
As a result, INNs have been used for 3D shape completion \cite{occflow, shapeflow, neural-part, cadex}, geometry processing \cite{yang2021geometry}, dynamic scenes reconstruction \cite{ndr}, and building animatable avatars with 3D scans \cite{ins}.
However, for video-based dynamic body deformation modeling, existing works only use non-invertible single-directional deformation.
In this work, we leverage the invertibility of the INNs to model the dynamic body motions and reconstruct high-quality dynamic body geometry.

\begin{figure*}[t]
\begin{center}
   \includegraphics[width=0.95\linewidth]{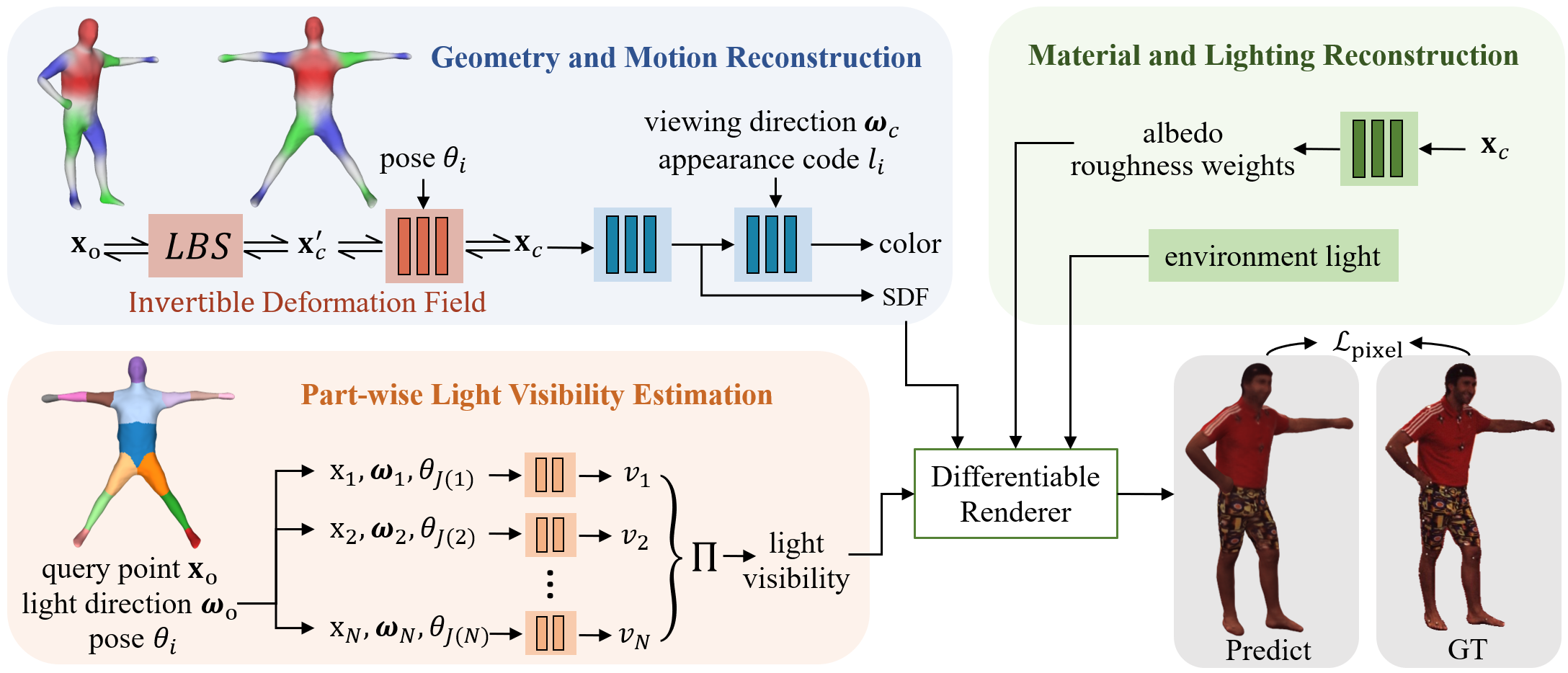}
\end{center}
\caption{The pipeline of our method. The invertible deformation field in \textit{Geometry and Motion Reconstruction} contributes to reconstruct more accurate dynamic body geometry (Sec.\ref{sec:geo_rec}). Then the networks in \textit{Part-wise Light Visibility Estimation} are trained to estimate pose-aware light visibility in an effective manner (Sec.\ref{sec:vis_est}). With these two parts fixed, the networks and lighting coefficients in \textit{Material and Light Estimation} are trained and optimized by the photometric losses (Sec.\ref{sec:mat_est}).}
\label{fig:pipeline}
\end{figure*}

\section{Method}

Given multi-view videos of a user with some arbitrary motions, our goal is to reconstruct a relightable and animatable avatar of the user.
The key challenge of this task is disentangling the geometry, material of the clothed body, and lighting, which is a highly ill-posed problem. 
To tackle this problem, we first reconstruct the body geometry from the input videos using the neural rendering techniques, where the geometry is modeled by a neural signed distance function (SDF) field and 
the dynamics of the human body are modeled with a rigid bone transformation of SMPL~\cite{loper2015smpl} model plusing an invertible neural deformation field (Sec.\ref{sec:geo_rec}, top left of Fig.\ref{fig:pipeline}).
Then, with the reconstructed geometry, we train a pose-aware part-wise light visibility estimation network, which is able to predict the light visibility of any query point under any light direction and body pose (Sec.\ref{sec:vis_est}, bottom left of Fig.\ref{fig:pipeline}).
Finally, with the visibility information, we achieve the disentangling of the material of the human body and the illumination parameters (Sec.\ref{sec:mat_est}, top right of Fig.\ref{fig:pipeline}).
Therefore, we can render a free-viewpoint video of the human with any target pose and illumination. 

\subsection{Geometry and Motion Reconstruction}
\label{sec:geo_rec}
Dynamic body deformation consists of articulated rigid motion and neural non-rigid deformation.
Correspondingly, we propose a mesh-based inverse skinning method and an invertible neural deformation field to map points between the canonical and the observation space bidirectionally.

\textbf{Mesh-based Inverse Skinning.}
The rigid motion is computed using linear blend skinning (LBS) algorithm \cite{pose_deformation}.
For point $\mathbf{x}_c$ in the canonical space, we use the bone transformation matrices $\{\mathbf{B}_b\}^{24}_{b=1}$ of the SMPL~\cite{loper2015smpl} model to transform $\mathbf{x}_c$ to $\mathbf{x}_o$ in the observation space (we omit the transformations of homogeneous coordinates for simplicity of notation):
\begin{equation}
    \mathbf{x}_o = \sum_{b=1}^{24} \left( w_b(\mathbf{x}_c) \mathbf{B}_b \right) \mathbf{x}_c
\end{equation}
where $w_b(\mathbf{x}_c)$ is the skinning weights of $\mathbf{x}_c$, and $\sum_{b=1}^{24} w_b(\mathbf{x}_c) = 1$.
Similarly, for $\mathbf{x}_o$ in the observation space, we can transform it back to the canonical space by:
\begin{equation}
    \mathbf{x}_c = \sum_{b=1}^{24} \left( w_b(\mathbf{x}_o) \mathbf{B}_b \right)^{-1} \mathbf{x}_o
\end{equation}
Similarly, for a query view direction $\mathbf{\omega}_o$ in the observation space, we can apply the same backward transformation to get the view direction $\mathbf{\omega}_c$ in the canonical space.

In volume rendering, we need to transform sampled ray points in the observation space to the canonical space (i.e. solve the inverse skinning problem) to query their SDF and color values.
However, determining the skinning weights of points in the observation space is non-trivial as $w_b$ is calculated in the canonical space rather than the observation space.
Many existing works, such as \cite{animatablenerf, peng2022animatable, humannerf}, rely on the posed SMPL mesh and use the skinning weights of neighboring SMPL mesh points to compute the inverse skinning weights of the ray points.
However, the naked SMPL mesh differs from the body surface, resulting in inaccurate weights.
Differently, we leverage the extracted explicit body mesh to compute the inverse skinning weights.
We first extract an explicit mesh of the body in the canonical space and compute the skinning weights of mesh vertices. 
Then, we use the LBS algorithm to deform the mesh to the observation space.
For any points in the observation space, we compute their skinning weights by the skinning weights of the nearest neighbor on the deformed body mesh.
As the deformed body mesh fits the actual body surface better than the naked SMPL mesh, our method does not suffer from the inaccuracies of skinning weights.

\textbf{Invertible Deformation Field.}
Since only rigid bone transformation is not enough for modeling the body motion, we use an invertible neural displacement field to model the non-rigid motions. 
As shown in Fig.\ref{fig:pipeline}, on the one hand, we apply non-rigid motion to the explicit mesh in canonical space.
On the other hand, for sampled ray points in observation space, we need to map them back to the canonical space.
Therefore, the neural displacement field should be able to transform points bidirectionally and ensure the cycle consistency of the transformation.
So, we involve an invertible neural network to represent the non-rigid motion. 
For a point $\mathbf{x} = \left[u, v, w\right]$ in the canonical space, we use the invertible network $D$ to apply displacement to it: 
\begin{equation}
    \mathbf{x}' = D(\mathbf{x}) = \left[u', v', w'\right]
\end{equation}
Besides, the invertible network $D$ can also transform $\mathbf{x}'$ back while keep the cycle consistency:
\begin{equation}
    \mathbf{x} =  D^{-1}(\mathbf{x}') = D^{-1}(D(\mathbf{x}))
\end{equation}
To keep the cycle consistency, we design a network similar to Real-NVP \cite{real-nvp}.
Specifically, we split the coordinates $[u, v, w]$ into two parts, for example, $[u, v]$ and $[w]$. 
During forward deformation, we assume the displacement of $[w]$ is decided by the value of $[u, v]$: 
\begin{equation}
    [w'] = [w] + f([u, v])
\end{equation}
and then the displacement of $[u, v]$ is decided by $[w']$:
\begin{equation}
    [u', v'] = [u, v] + g([w'])
\end{equation}
With this two-step forward deformation $D$, we can directly get an invertible backward deformation $D^{-1}$ which deforms point $[u', v', w']$ to $[u, v, w]$ as follows:
\begin{equation}
\begin{aligned}
    [u, v] & = [u', v'] - g([w']) \\
    [w] & = [w'] - f([u, v])
\end{aligned}
\end{equation}
The functions $f(\cdot), g(\cdot)$ are implemented as MLPs, they form a transformation block of the invertible network $D$.
As the aforementioned $f(\cdot)$ makes the deformation decided by $[u, v]$ only, we stack more transformation blocks and change the split of $[u, v, w]$ in these blocks. 
We assume that the non-rigid deformations are pose-dependent, for the $i$th frame, we use the body pose $\theta_i$ as the condition of the network $D$. 
Besides, we found that it is hard to learn the deformation using only the pose and coordinates as conditions.
Thus, we use the skinning weights of the query points $\mathbf{W}(\mathbf{x})\ \in \mathbb{R}^{24}$ as an additional condition, which leads to better results.
The displacement field $D$ can be formulated as $\mathbf{x}' = D(\mathbf{x}, \theta_i, \mathbf{W}(\mathbf{x}))$.
The use of skinning weights will slightly affect the cycle consistency of the deformation network, as $\mathbf{W}(\mathbf{x})$ is not strictly equal to $\mathbf{W}(\mathbf{x}')$.
But we found the skinning weights field in the canonical pose is smooth, and the deformations are relatively small, so they are almost the same, the sacrifice on cycle consistency is negligible.

\textbf{Network Training.}
To supervise these neural fields with videos, we use the technique proposed in VolSDF \cite{volsdf} to convert the SDF values to density and conduct volume rendering. 
For the color field, we introduce learnable per-frame appearance latent codes $\{l_i\}_{i=1}^N$ to model the dynamic appearance, where $N$ is the number of frames.
Besides, we optimize the pose vectors $\{\theta_i\}_{i=1}^N$, as the initial poses may not be accurate.
In sum, the training parameters contain the SDF network, the color network, the deformation network $D$, the appearance latent codes $\{l_i\}_{i=1}^N$ and the pose vectors $\{\theta_i\}_{i=1}^N$. 
The training loss consists of rendering photometric loss and multiple regularizers:
\begin{equation}
    \mathcal{L} = \lambda_{\text{pixel}}\mathcal{L}_{\text{pixel}} + \lambda_{\text{mask}}\mathcal{L}_{\text{mask}} + \lambda_{\text{eik}}\mathcal{L}_{{\text{eik}}} + \lambda_{\text{disp}}\mathcal{L}_{{\text{disp}}}
\end{equation}
where $\mathcal{L}_{\text{pixel}}$ is an L2 pixel loss for predicted color, $\mathcal{L}_{\text{mask}}$ is a binary cross-entropy loss for the rendering object mask and input mask, $\mathcal{L}_{{\text{eik}}}$ is the Eikonal regularization term \cite{eik}, $\mathcal{L}_{{\text{disp}}}$ is an L2 regularizer for the output displacements.
For more details about network architecture and training, please refer to the supplemental document.

\subsection{Part-wise Light Visibility Estimation}
\label{sec:vis_est}
With the reconstructed geometry, we then conduct pose-aware light visibility estimation.
Modeling the visibility allows for the extraction or generation of shadows on images, which helps to better disentangle material and lighting from input images as well as produce physically plausible shadow effects in rendered images.
Given a query point $\mathbf{x}$ and a query light direction $\mathbf{\omega}$, our goal is to train a network to predict whether the query point will be lighted or occluded by the body in a certain a pose and light direction.

Traditionally, estimating light visibility is solved by performing ray tracing.
However, for implicit neural network-based methods, tracing a path of light requires numerous queries, as we need to trace all possible lighting directions for one 3D point, which is very time-consuming.
Thus, existing methods \cite{NeRFactor, InvRender, Relighting4D} use MLPs to re-parameterize and speed up this process as $V\left(\mathbf{x}, \mathbf{\omega}\right) \mapsto v$, where $v=1$ indicates the point is visible to the light from $\omega$ direction.
However, with the motion of the human body, light visibility changes dramatically.
Relighting4D \cite{Relighting4D} leverages temporally-varying latent codes to model these changes, but it is limited to seen poses as there is no latent code for unseen motions.
To solve this problem, we need to make it pose-aware for light visibility estimation.
A naive approach is to use the pose vectors as the condition of the visibility network, but we found this approach does not work well as the relationship among pose, lighting, and shadow is too complex to be modeled.

Our observation is that how light rays are blocked is determined by the object geometry, even though the human body as a whole can be in different complex shapes caused by pose changes, for a single body part, its geometry changes are relatively small among different poses.
So, we divide the human body into $N(=15)$ parts as shown in the orange rectangle in Fig.\ref{fig:pipeline}, where different colors denote different body parts.
Then for each body part, we train a neural network respectively to predict how the body part blocks the lights.
Finally, we combine the light visibility of all body parts by multiplying all the predicted visibility.
Thus, our method achieves light visibility prediction of any query points, light directions, and body poses.

To be specific, given the query point $\mathbf{x}_o$ and light direction $\mathbf{\omega}_o$ in observation space, we first transform them to the local coordinate of each body part:
\begin{equation}
    \mathbf{x}_i = \mathbf{B}_i^{-1} \mathbf{x}_o, \quad \mathbf{\omega}_i = \mathbf{B}_i^{-1} \mathbf{\omega}_o
\end{equation}
where $\mathbf{B}_i$ is the bone transformation of the $i$th body part.
Besides, although the geometry changes of body parts are relatively small, there are still some pose-dependent deformations. 
And the geometry of a body part is majorly affected by the poses of its neighboring joints, so we use them as the networks' condition.
We denote the neighboring joints of body part $i$ as $J(i)$.
So, the visibility network of a body part $i$ can be formulated as:
\begin{equation}
    V_i\left(\mathbf{x}_{i}, \mathbf{\omega}_{i}, \mathbf{\theta}_{J(i)}\right) \mapsto v_i
\end{equation}

For network training, we sample different query points, light directions, and body poses, then we perform ray tracing to compute the ground truth light visibility of each body part.
We impose binary cross-entropy loss to train the networks for visibility estimation.

\subsection{Material and Light Estimation}
\label{sec:mat_est}
At this stage, we fix the geometry and light visibility estimation modules and optimize the material network and light parameters as shown in the green rectangle in Fig.\ref{fig:pipeline}.
Here, we parameterize the material using the Disney BRDF \cite{disneyBRDF} model and use albedo and roughness to represent the material.
However, we found that directly optimizing the roughness is difficult.
Similar to \cite{zhuo17, li22, psnerf}, we use a weighted combination of specular bases.
Each basis is defined by a different roughness value.
For a query point in the canonical space, we use an implicit neural network $M$ to predict its albedo value and roughness weights.
For environment light, we parameterize it using $L=128$ spherical Gaussians (SGs) \cite{SG}:
\begin{equation}
    E\left(\mathbf{\omega}_{i}\right)=\sum_{j=1}^L G\left(\mathbf{\omega}_{i} ; \mathbf{\xi}_j, \lambda_j, \mathbf{\mu}_j\right)
\end{equation}
where $\mathbf{\omega}_i\in \mathbb{S}^2$ is the query lighting direction, $\mathbf{\xi}_j\in \mathbb{S}^2$ is the lobe axis, $\lambda_j \in \mathbb{R}_{+}$ is the lobe sharpness, $\mu_j \in \mathbb{R}^3$ is the lobe amplitude. 
To compute the visibility of an SG light, we sample 4 directions around the lobe axis based on the distribution defined by $\lambda_j$. 
Then we predict their visibilities by the trained light visibility estimation network and use the weighted sum of these samples as the visibility of the SG light.

With geometry, material, environment light, and light visibility, we can render images of the human body using a differentiable renderer.
The rendering equation computes the outgoing radiance $L_o$ at point $\mathbf{x}$ viewed from $\mathbf{\omega}_o$:
\begin{equation}
    L_o\left(\mathbf{x}, \mathbf{\omega}_o\right)=\int_{\Omega} L_{i }\left(\mathbf{x}, \mathbf{\omega}_i\right) R\left(\mathbf{x}, \mathbf{\omega}_i, \mathbf{\omega}_o, \mathbf{n}\right)\left(\mathbf{\omega}_i \cdot \mathbf{n}\right) \mathrm{d} \mathbf{\omega}_i
\end{equation}
where $L_{i}\left(\mathbf{x}, \mathbf{\omega}_i\right)$ is the incident radiance of point $\mathbf{x}$ from direction $\mathbf{\omega}_i$, which is determined by the environment light $E$ and masked by the light visibility. 
$R\left(\mathbf{x}, \mathbf{\omega}_i, \mathbf{\omega}_o, \mathbf{n}\right)$ is the Bidirectional Reflectance Distribution Function (BRDF) which is determined by the albedo values and roughness weights predicted by $M$. 

To train the material network $M$ and the light parameters of $E$, we use L1 pixel loss between the rendered images and the recorded images.
However, there are strong ambiguities in solving material and lighting, so we apply some regularization strategies.
First, the material network is designed as an encoder-decoder architecture following \cite{InvRender}, so that we can impose constraints on the latent space to ensure the sparsity of albedo and roughness weights. 
We denote the encoder and decoder of $M$ as $M_E$ and $M_D$, for a query point $\mathbf{x}$ in the canonical space, its latent vector is $\mathbf{z} = M_E(\mathbf{x}) \in \mathbb{R}^N$. 
For $K$ latent codes in a batch $\{\mathbf{z}_i\}_{i=0}^K$, we impose Kullback-Leibler divergence loss to encourage the sparsity of the latent space:
\begin{equation}
    \mathcal{L}_{\text{kl}} = \sum_{j=1}^N \mathrm{KL}\left(\rho || \hat{\rho}_j\right)
\end{equation}
where $\hat{\rho}_j$ is the average of the $j$th channel of $\{\mathbf{z}_i\}_{i=0}^K$, $\rho$ is set to 0.05.
Moreover, we apply smooth loss to both the latent vectors and the output albedo and roughness weights by adding perturbations:
\begin{equation}
\begin{aligned}
    \mathcal{L}_{\text{smooth}} = & \lambda_\mathbf{z} \|M_D(\mathbf{z}) - M_D(\mathbf{z} + \mathbf{\xi}_\mathbf{z})\|_1 + \\ 
    & \lambda_\mathbf{x} \| M(\mathbf{x}) - M(\mathbf{x} + \mathbf{\xi}_\mathbf{x})\|_1
\end{aligned}
\label{eq:lsmooth}
\end{equation}
where $\mathbf{\xi}_\mathbf{z}$ and $\mathbf{\xi}_\mathbf{x}$ are the perturbations of the latent code $\mathbf{z}$ and the query point $\mathbf{x}$, which is sampled from a Gaussian distribution with zero mean and 0.01 variance.

In sum, the full training loss for this stage is:
\begin{equation}
    \mathcal{L} = \lambda_{\text{pixel}} \mathcal{L}_{\text{pixel}} + \lambda_{\text{kl}} \mathcal{L}_{\text{kl}} + \lambda_{\text{smooth}} \mathcal{L}_{\text{smooth}}
    \label{eq:lmat}
\end{equation}

With the trained geometry field, the deformation field, the light visibility estimation networks $V$, and the material network $M$, we can render the avatar in novel poses, lightings, and viewpoints. 
Thus we achieve a relightable and animatable neural avatar.

\begin{table*}[t]
\begin{center}
\scalebox{0.9}{
\begin{tabular}{cccccccccc}
\hline
\multirow{2}{*}{Method} & \multicolumn{3}{c}{Albedo Map} & \multicolumn{3}{c}{Relighting (Training poses)} & \multicolumn{3}{c}{Relighting (Novel poses)} \\
                        & PSNR$\uparrow$ & SSIM$\uparrow$ & LPIPS$\downarrow$ & PSNR$\uparrow$ & SSIM$\uparrow$ & LPIPS$\downarrow$ & PSNR$\uparrow$ & SSIM$\uparrow$ & LPIPS$\downarrow$  \\
\hline
Relighting4D~\cite{Relighting4D} & 21.5103 & 0.8320 & 0.2299 & 19.7323 & 0.7568 & 0.2721 & 16.7475 & 0.6729 & 0.3330 \\
\hline
Ours w/o visibility & 24.7611 & 0.8918 & 0.1655 & 23.7758 & 0.8376 & 0.2223 & 18.9768 & 0.7333 & 0.2638 \\
Ours w/o part-wise visibility & \textbf{25.2150} & \textbf{0.8921} & 0.1652 & 24.7064 & 0.8462 & 0.2173 & 19.7119 & 0.7452 & 0.2580 \\
Ours & 25.1666 & 0.8919 & \textbf{0.1645} & \textbf{25.3477} & \textbf{0.8546} & \textbf{0.2124} & \textbf{19.8622} & \textbf{0.7518} & \textbf{0.2533} \\
\hline
\end{tabular}
}
\caption{Quantitative comparison of the reconstructed albedo and the relighting results on synthetic data.}
\label{tab:syn}
\end{center}
\end{table*}

\begin{figure}[t]
\begin{center}
   \includegraphics[width=1.0\linewidth]{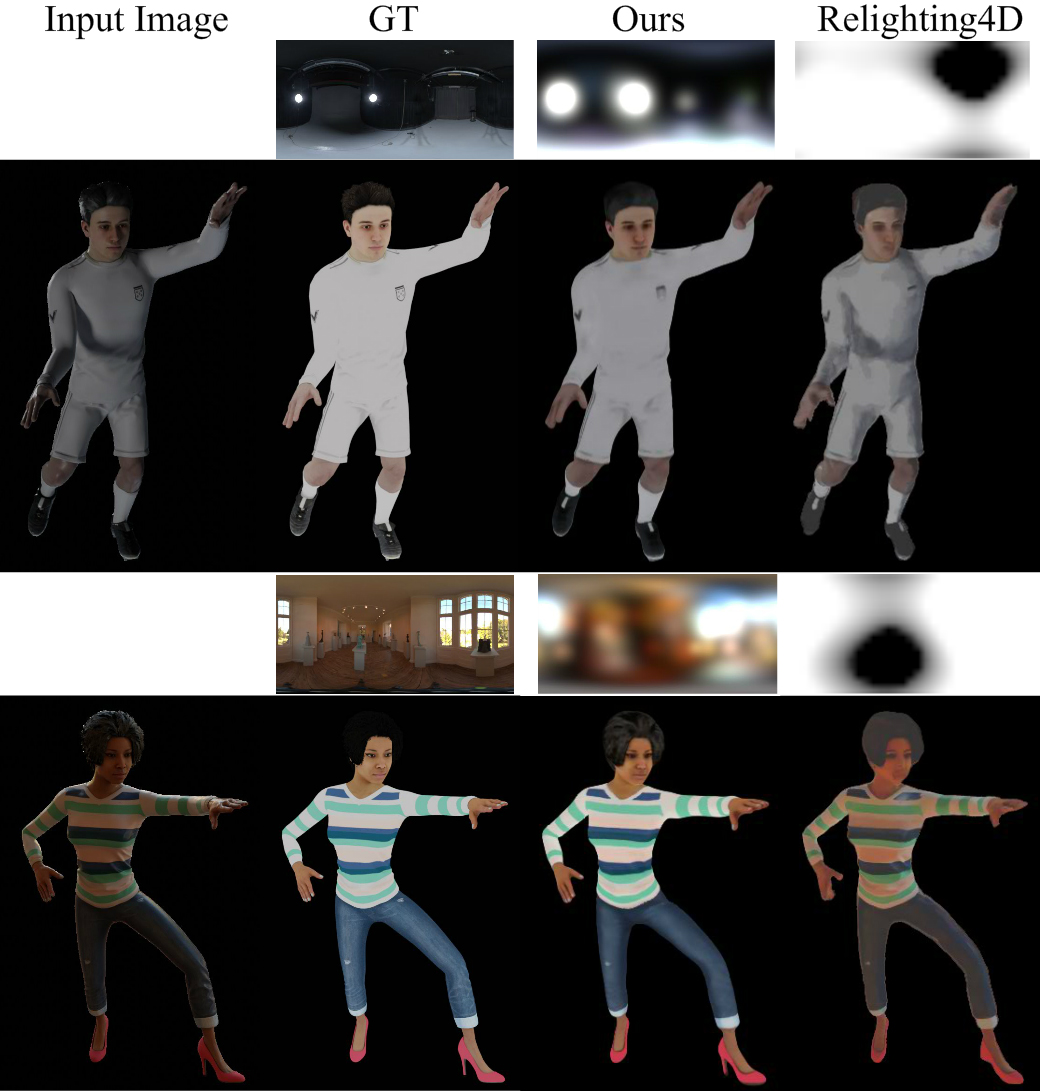}
\end{center}
\caption{Qualitative comparison of the reconstructed albedo and lighting on synthetic data. Environment lighting is shown on top of the albedo in each result.}
\label{fig:albedo_rec}
\end{figure}

\begin{figure}[t]
\begin{center}
   \includegraphics[width=1.0\linewidth]{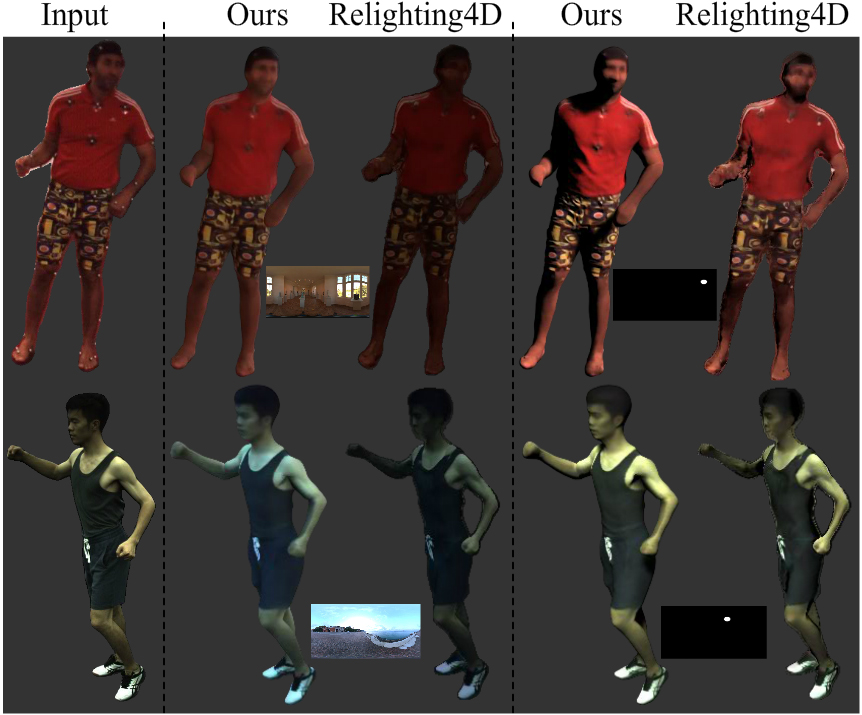}
\end{center}
\caption{Qualitative comparison of relighting results on real data. The environment lighting of the rendered results is shown at the bottom.}
\label{fig:relit_real}
\end{figure}

\section{Experiments}

In this section, we evaluate the performance of our method qualitatively and quantitatively.
First, we introduce the used datasets.  
Then we compare our method with the state-of-the-art human relighting method Relighting4D \cite{Relighting4D}.
Since our geometry reconstruction is improved by the proposed invertible deformation field, we also compare our method with the state-of-the-art video-based human geometry reconstruction methods ARAH \cite{ARAH} and \citet{peng2022animatable}.
Next, we perform ablation studies to validate our key design choices.
Finally, we show the synthesized results on various characters with various body motions under various lightings, and video results can be seen in the supplemental video.

\subsection{Datasets}

We use both real and synthetic datasets for comparisons and evaluations.
For the real dataset, we use multi-view dynamic human datasets including the ZJU-MoCap \cite{neuralbody}, Human3.6M \cite{h36m}, DeepCap \cite{deepcap} and PeopleSnapshot \cite{alldieck2018video} dataset.
To perform a quantitative evaluation, we create a new synthetic dataset.
We leverage 4 rigged characters from Mixamo\footnote{\url{https://www.mixamo.com/}} and transfer the body motion from the ZJU-MoCap dataset to generate motion sequences. Each sequence contains 100 frames.
Then, we use Blender\footnote{\url{https://www.blender.org/}} to render multi-view videos under different illuminations with HDRI environment maps from PolyHaven\footnote{\url{https://polyhaven.com/}}.
Besides, we use 4 OLAT light sources for relighting evaluations.

\subsection{Comparisons}
Since Relighting4D \cite{Relighting4D} is the state-of-the-art for video-based human motion relighting, we compare our full method with it on albedo estimation, lighting reconstruction, and relighting under training/novel poses.
Body geometry is an intermediate result of our method, we also compare it with the state-of-the-art video-based human geometry reconstruction methods \citet{peng2022animatable} and ARAH \cite{ARAH}.

\textbf{Material Estimation and Relighting.}
The comparison results with Relighting4D \cite{Relighting4D} are shown in Fig.\ref{fig:albedo_rec}.
Relighting4D cannot disentangle the lighting and appearance very well, as we can see noticeable errors on both the estimated environment map and the reconstructed albedo.
For example, there are shadows wrongly baked into albedo in the result on the top right side.
Besides, some lighting information is backed into albedo in the result on the bottom right side.
For numerical comparisons, we use Peak Signal-to-Noise Ratio (PSNR), Structural Similarity Index Measure (SSIM) \cite{ssim}, and Learned Perceptual Image Patch Similarity (LPIPS) \cite{lpips} as metrics.
In Tab.\ref{tab:syn}, we show the numerical albedo estimation result on synthetic data, which also indicates our improvement in albedo estimation.
Tab.\ref{tab:syn} also shows the final relighting results for both training poses and novel poses, both show that we achieve noticeably better results than Relighitng4D.
Note that for novel poses, there are misalignments in the geometry between the animated geometry and the ground truth geometry, which leads to noticeable performance drops for novel poses.
Besides, qualitative results for relighting on real datasets are shown in Fig.\ref{fig:relit_real} (please zoom in for better comparison).
The overall lighting effect is better rendered by our method as shown on the left, and the spatial variant effect caused by point light is also correctly generated by our method as shown on the right due to the success of visibility modeling. More video results can be seen in our supplementary video.
Notice that as Relighting4D relies on per-frame latent codes to model the dynamics, it does not support novel poses synthesis by design. So, when performing relighting for a novel pose, we find the closest pose in its training poses and use its latent code for inference.

\textbf{Geometry Reconstruction.}
We evaluate different methods on our synthetic dataset with ground truth geometry and use point-to-surface distance (P2S) and Chamfer Distance (CD) as metrics.
The results as shown in Tab.\ref{tab:geo}, our method outperforms the compared two methods on all test sequences. 
We also show qualitative comparisons of rendering images in the real dataset in Fig.\ref{fig:geo_artifact}.
We can find that there are obvious artifacts in the results of \citet{peng2022animatable} and ARAH in the elbow and hand regions.
While the result of our method does not suffer from the artifacts, as our mesh-based inverse skinning helps to find accurate correspondences between the observation space and the canonical space.
In contrast, \citet{peng2022animatable} use posed SMPL models to compute the backward skinning weights which leads to worse correspondences, especially for regions with body contacts.
ARAH involves iterative root-finding to compute the correspondences, but the optimization sometimes fails to converge, thus also leading to artifacts.

\begin{table}[]
\begin{center}
\scalebox{0.85}{
\begin{tabular}{ccccccc}
\hline
                     & Method      & S1    & S2    & S3    & S4    & Avg \\ \hline
\multirow{5}{*}{P2S$\downarrow$} & Peng et al. & 0.387 & 0.359 & 0.339 & 0.339 & 0.356   \\
                     & ARAH        & 0.317 & 0.340 & 0.325 & 0.280 & 0.316   \\
                     & Ours w/o MIS  & 0.241 & 0.230 & 0.234 & 0.241 & 0.237   \\
                     & Ours w/o $\mathbf{W}$ & 0.246 & 0.247 & 0.243 & 0.256 & 0.248   \\
                     & Ours        & \textbf{0.185} & \textbf{0.179} & \textbf{0.182} & \textbf{0.184} & \textbf{0.182}   \\ \hline
\multirow{5}{*}{CD$\downarrow$}  & Peng et al. & 0.656 & 0.864 & 0.521 & 0.528 & 0.642   \\
                     & ARAH        & 0.531 & 0.714 & 0.477 & 0.441 & 0.541   \\
                     & Ours w/o MIS & 0.462 & 0.666 & 0.423 & 0.391 & 0.485   \\
                     & Ours w/o $\mathbf{W}$ & 0.479 & 0.700 & 0.427 & 0.424 & 0.507   \\
                     & Ours        & \textbf{0.395} & \textbf{0.609} & \textbf{0.358} & \textbf{0.363} & \textbf{0.431}   \\ 
\hline
\end{tabular}
}
\caption{Quantitative comparison of the reconstructed geometry on synthetic data.}
\label{tab:geo}
\end{center}
\end{table}

\subsection{Ablation Study}
\label{sec:ablation}
Here, we evaluate our two key components: mesh-based inverse skinning (MIS) and part-wise visibility estimation. 
The MIS based on the invertible deformation makes it possible to deform the more accurate mesh in the canonical space to the observation space to calculate skinning weights. 
Otherwise, the naked SMPL mesh with large geometry errors has to be used. 
So, we compare our method with using the SMPL mesh in weights calculation. 
Besides, using the part-wise design achieves accurate light visibility estimation, which is crucial to generate self-occlusion effects on bodies.
To evaluate it, we compare it with two alternatives, removing the light visibility module and using only one neural network to predict the light visibility.

\begin{figure}[t]
\begin{center}
   \includegraphics[width=1.0\linewidth]{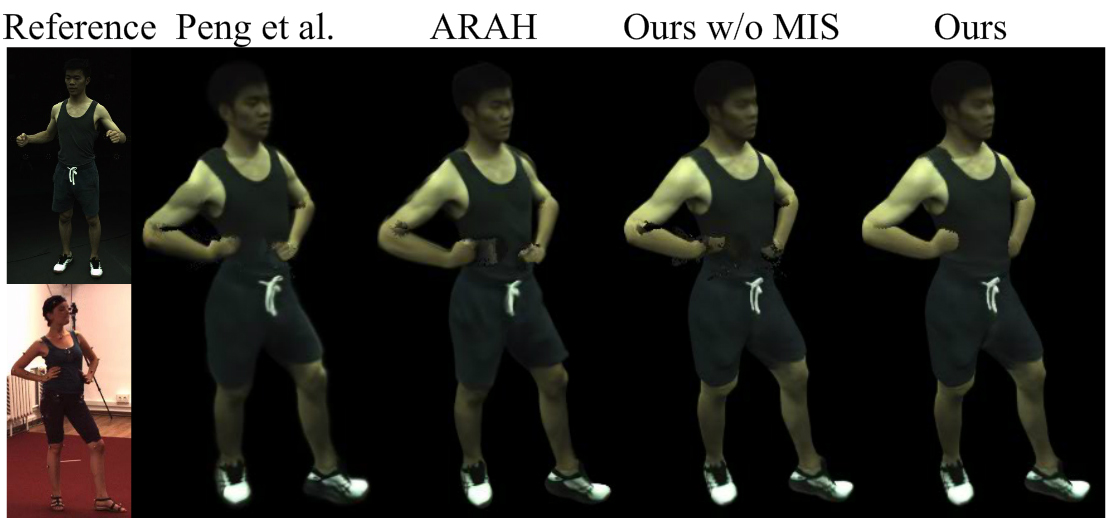}
\end{center}
\caption{Qualitative results of novel poses synthesis on real data. This novel pose results reflect the accuracy of the reconstructed geometry to a certain extent. }
\label{fig:geo_artifact}
\end{figure}

\textbf{Mesh-based Inverse Skinning.}
As shown in Tab.\ref{tab:geo}, the reconstruction errors without the mesh-based inverse skinning are consistently larger.
We also show the qualitative result in Fig.\ref{fig:geo_artifact}, using SMPL mesh to compute the skinning weights leads to artifacts in the contact regions.

Besides, we evaluate the effect of the condition of skinning weights $\mathbf{W}$ (in Sec.~\ref{sec:geo_rec}) in the invertible deformation network.
As shown in Tab.\ref{tab:geo}, we can also find that removing the condition of $\mathbf{W}$ also leads to worse results.

\begin{figure}[t]
\begin{center}
   \includegraphics[width=1.0\linewidth]{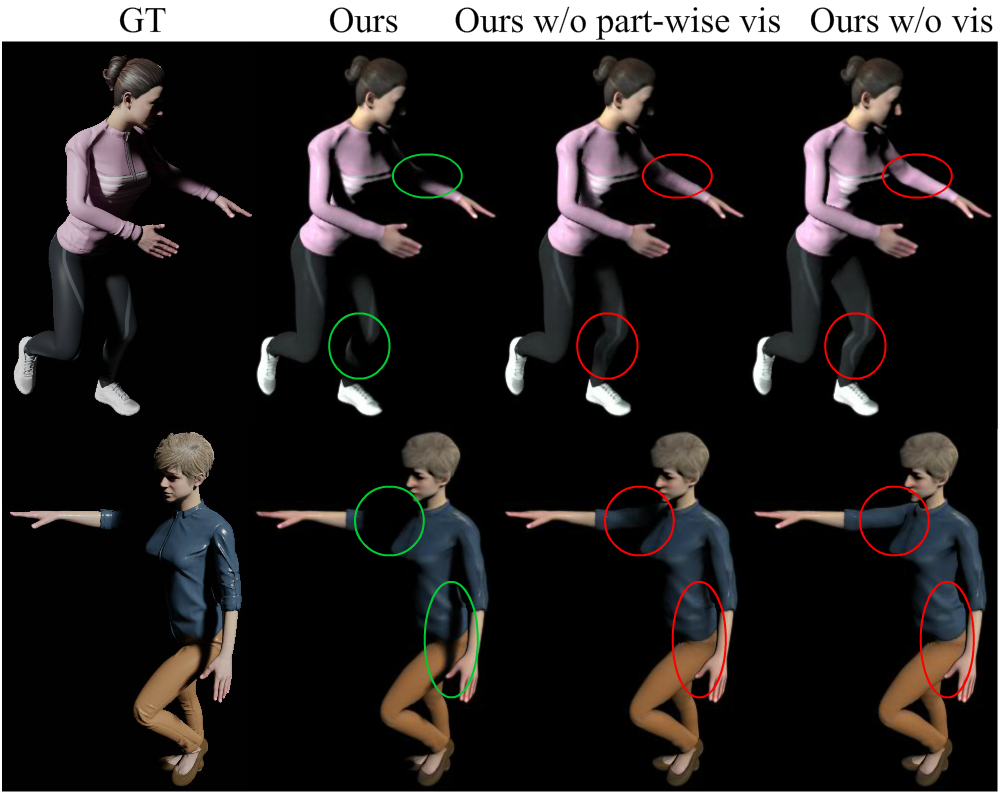}
\end{center}
\caption{Ablation study on part-wise light visibility. See our method synthesizing plausible self-occlusions.}
\label{fig:ablation_lvis}
\end{figure}

\textbf{Part-wise Visibility Estimation.}
We show quantitative comparisons in Tab.\ref{tab:syn}, the results show that although the albedo map reconstruction qualities are similar, our method with part-wise visibility estimation achieves the best results on relighting.
Furthermore, We show qualitative results in Fig.\ref{fig:ablation_lvis}.
We can see that without light visibility modeling (results in the fourth column), the self-occlusion effect cannot be generated at all.
With the baseline light visibility modeling, self-occlusion can be partly generated for some poses (results of the third column).
For our final solution, the received lighting for different body regions on the novel poses are well modeled and thus the relighting results are consistent with the ground truth rendering.

\section{Limitations}
The network training takes about 2.5 days in total on a single RTX 3090 GPU, and it takes about 40 seconds to render an image with a resolution of $512\times 512$ during inference (more details in the supplemental document).
Integrating instant training techniques like Instant-NGP \cite{instant-ngp} may improve the efficiency of our technique.
It is still hard for our method to animate pose-dependent wrinkle deformations (especially for loose clothing) or generate global illumination effects, which are also open problems in this topic.
Our method only considers the body motion rather than the face and hands, while recent works~\cite{zheng2023avatarrex, shen2023xavatar} provide possibilities to handle them.

\section{Conclusion}
This is the first work that reconstructs relightable and animatable neural avatars with plausible shadow effects from sparse human videos.
For dynamic body geometry modeling, the proposed invertible deformation field provides a novel and effective way to solve the inverse skinning problem.
Besides, the part-wise light visibility modeling solves the problem of dynamic object relighting based on neural fields.
Benefiting from the two techniques, our method succeeds in disentangling the geometry, material of the clothed body, and lighting, thus building a relightable and animatable neural avatar in a lightweight setting.

\section{Acknowledgments}

This work was supported by the National Key R\&D Program of China (2018YFA0704000), the NSFC (No.62021002), and the Key Research and Development Project of Tibet Autonomous Region (XZ202101ZY0019G). This work was also supported by THUIBCS, Tsinghua University, and BLBCI, Beijing Municipal Education Commission. Jun-Hai Yong and Feng Xu are the corresponding authors.

\bibliography{main}

\clearpage
\appendix
\section{Overview}

In the appendix, we first present the implementation details of our method, including the network architectures, the training process, and the used datasets. 
Second, we evaluate of the cycle consistency of the deformation field.
Then, we show the results of the visualization of the correspondences.
Next, we compare our method with state-of-the-art non-relightable neural avatar methods.
Finally, we show additional results of our method in different datasets to demonstrate its effectiveness.

\section{Implementation Details}

\begin{figure*}[t]
\begin{center}
   \includegraphics[width=0.7\linewidth]{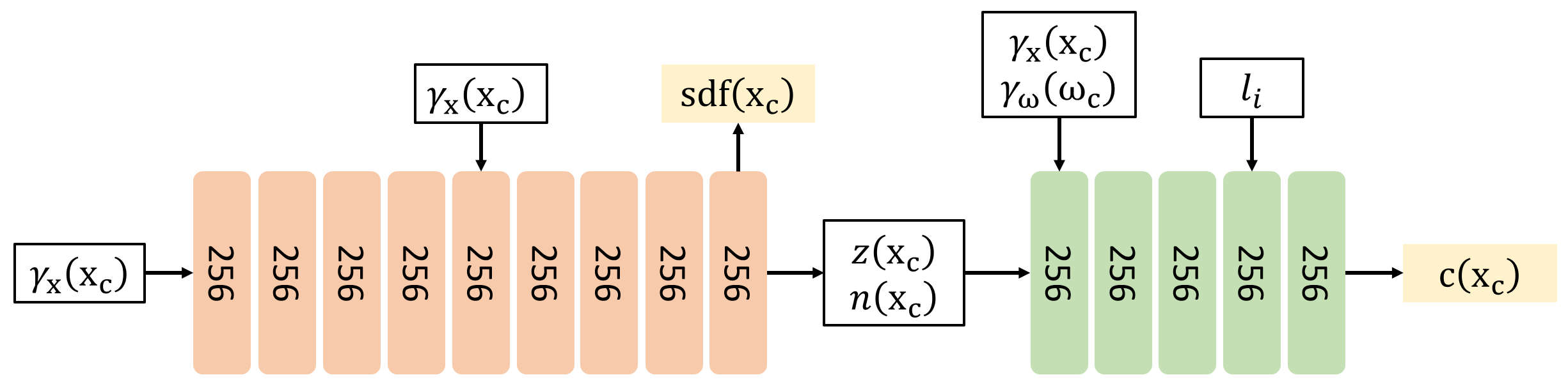}
\end{center}
\caption{Architecture of the geometry and color network.}
\label{fig:geo_net}
\end{figure*}

\begin{figure*}[t]
\begin{center}
   \includegraphics[width=0.85\linewidth]{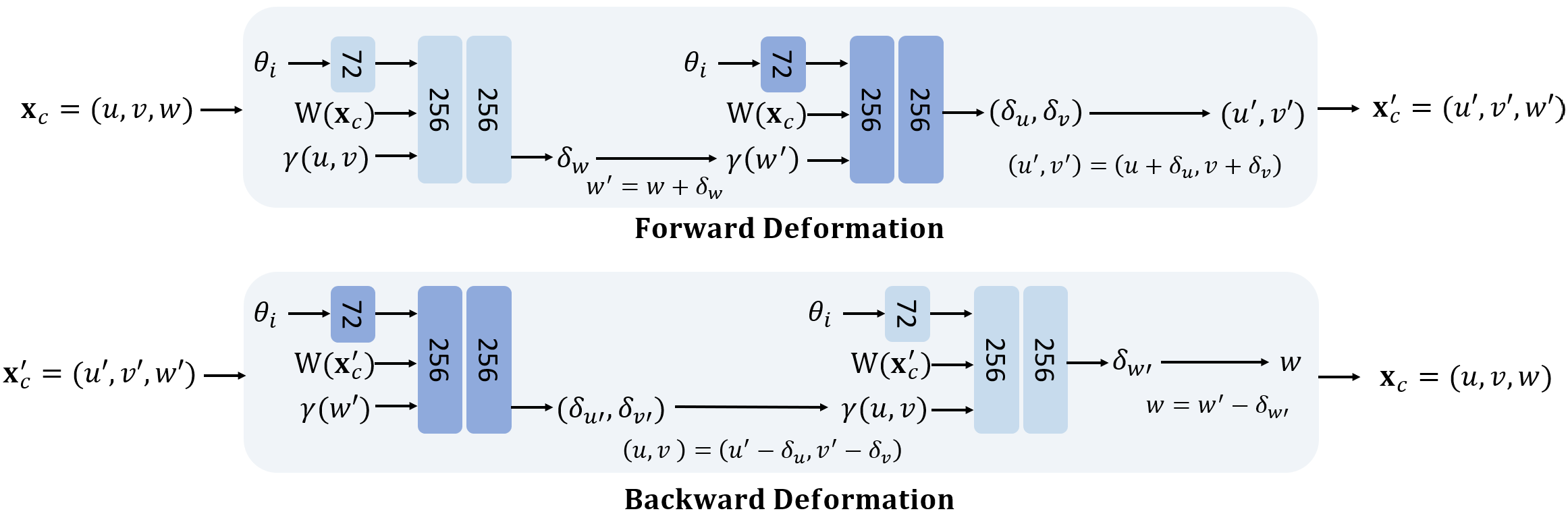}
\end{center}
\caption{Architecture of the basic block in the invertible deformation network.}
\label{fig:def_net}
\end{figure*}

In this section, we provide more details about the network architectures, the training process and the used datasets.

\subsection{Network Architectures}

First, we show the structure of the geometry and color network used in the geometry and motion reconstruction stage, illustrated in Fig.~\ref{fig:geo_net}.
The orange part is the geometry network, the green part is the color network.
The $\gamma_\mathbf{x}(\cdot), \gamma_\mathbf{\omega}(\cdot)$ are the positional encoding functions for query points and view directions. 
The frequencies are 10 for positions, and 4 for directions, which is the same for all the neural networks in our method.
In Fig.~\ref{fig:geo_net}, $\mathbf{z}(\mathbf{x}_c)$ is the feature vector with a size of 256, $\mathbf{n}(\mathbf{x}_c)$ is the normal of $\mathbf{x}_c$, calculated through the gradient of the SDF field. Additionally, $l_i$ is the appearance latent code of the $i$th frame with a size of 128. 

Next, we show the basic transformation block of the invertible displacement network in Fig.~\ref{fig:def_net}, with the procedures for both forward and backward deformation. 
$\mathbf{W}(\mathbf{x}_c) \in \mathbb{R}^{24}$ is the skinning weights vector, $\gamma(\cdot)$ is the positional encoding function.
The full displacement network consists of 3 blocks, and $(u, v, w)$ are split in 3 different orders.

In the light visibility estimation module, each sub-network is an MLP with 3 hidden layers and a width of 128.
In the ablation study, the light visibility estimation network without part-wise design is an MLP with 4 hidden layers and 256 in width. It should be noted that the size of this network is the same as the light visibility network of Relighting4D~\cite{Relighting4D}.

In the material network $M$, the encoder is an MLP with 4 hidden layers and a width of 512, the dimension of the latent space is 32, and the decoder is an MLP with 2 hidden layers and a width of 128.
The roughness weights in $M$ contain 9 different specular bases defined by different roughness values. 

\subsection{Training Details}

For geometry and motion reconstruction, the loss weights are $\lambda_{\text{pixel}}=1, \lambda_{\text{mask}}=1, \lambda_{\text{eik}}=0.01, \lambda_{\text{disp}}=0.02$.
The $\mathcal{L}_{\text{mask}}$ term is implemented in a coarse to fine manner as in IDR \cite{idr}.
For a sampled ray $r$, we find the minimum SDF value $s_r$ along the ray and apply binary-cross-entropy loss as follow:
\begin{equation}
    \mathcal{L}_{\text{mask}} = \sum_{r\in \mathcal{R}} \text{BCE}\left(\text{sigmoid}(-\alpha s_r), M(r)\right)
\end{equation}
where $M(r) \in \{0, 1\}$ is the ground truth mask for the ray $r$, $\alpha$ is initially set to 50 and multiplied by a factor of 2 every 20 epochs. The number of multiplications of $\alpha$ is up to 5.

At the geometry and motion reconstruction stage, we train the network for 400 epochs. For the first 40 epochs, we use the posed SMPL \cite{loper2015smpl} mesh to compute the skinning weights in the observation space, and the output displacements are set to zero.
After this warm-up process, we replace the SMPL mesh with the extract body mesh, and use the displacement network to apply bidirectional deformations.
The body mesh is extracted from the canonical SDF field using marching cubes algorithm \cite{marchingcubes}.

For the training of the light visibility network, we sample 2000 poses from the AIST++ \cite{aist} dataset.
For each pose, we randomly sample 4 light directions and 16,384 3D points and compute their light visibility.
The networks are trained for 32 epochs.
Please note that the training and novel poses in Tab.~1 of the main paper are both unseen for the light visibility network. 
The training poses in the table are the poses seen in the input videos, and the novel poses are just used to evaluate the animatable capability.

At the material and lighting optimization stage, the loss weights are $\lambda_{\text{pixel}}=1, \lambda_{\text{kl}}=0.001, \lambda_{\text{smooth}}=0.01, \lambda_{\mathbf{z}}=1.0, \lambda_{\mathbf{x}}=0.05$.
The models are trained for 200 epochs.

During volume rendering, we first uniformly sample 256 points per ray.
For the geometry and motion reconstruction stage, we only keep the ray points close to the body mesh for color integration.
For the material and light estimation stage, as the trained geometry is fixed, the material network only needs to query the sampled points with non-zero weights for volumetric rendering.

We implement our model using PyTorch and use the Adam \cite{adam} optimizer for training.
The learning rate is $5e-4$ for the first and third stages and $1e-3$ for the second stage.
All the models are trained on a single NVIDIA RTX 3090 GPU.
It takes about a day each to train the first and third stages, and the light visibility estimation network takes approximately 12 hours to train.

During inference, our method with the part-wise visibility network takes about 40s to render an image of $512\times512$ resolution. 
Besides, it takes about 18s and 22s for our method without the visibility network and with a single visibility network respectively. 
The additional time for querying 15 body parts for light visibility is acceptable and the part-wise light visibility module achieves significantly better results than the single network.
In contrast, computing light visibility through ray tracing takes about 430s per frame, causing the training of the network to be more than 10 days. 

\subsection{Dataset Details}
Here we provide more details about the used datasets.
For the synthetic dataset, we synthesize 4 body motion sequences as the training set, each containing 100 frames.
We sample 10 frames evenly from each sequence for evaluation.
For the novel pose evaluation, we sample 10 out-of-distribution poses from the AIST++ \cite{aist} dataset.
The videos in the dataset are generated under 8 different views, with 4 views used for training and the other 4 views used for evaluation. 

For evaluation on real datasets, we follow the data preprocessing method and the training-test split of \citet{peng2022animatable} for the ZJU-MoCap \cite{neuralbody}, Human3.6M \cite{h36m}, DeepCap \cite{deepcap} and PeopleSnapshot \cite{alldieck2018video} datasets.
In our experiments, there are 4 input views used in the ZJU-MoCap and DeepCap datasets, 3 input views used in the Human3.6M dataset, and a single input view used in the PeopleSnapshot dataset. 
In the ZJU-MoCap, DeepCap and PeopleSnapshot datasets, the number of frames in the input videos is 300.
In the Human3.6M dataset, the number of training frames ranged from 150 to 260.

\section{Evaluation of the Deformation Cycle Consistency}

We propose an invertible deformation field that allows for bidirectional deformation while maintaining cycle consistency.
For ordinary invertible neural networks, the cycle consistency is strictly satisfied.
However, we involve the skinning weights $\mathbf{W}(\mathbf{x})$ as an additional condition of the deformation network to improve the geometry reconstruction results, which slightly sacrifices the cycle consistency of the deformation field.
In this section, we evaluate the cycle consistency of the proposed invertible deformation network.

To evaluate the cycle consistency, we first sample some points on the body mesh.
Then for a sampled point $\mathbf{x}$, we apply the forward displacement $f(\mathbf{x})$ to obtain the deformed point $\mathbf{x}' = \mathbf{x} + f(\mathbf{x})$.  
Then, we apply the backward displacement $f^{-1}(\mathbf{x}')$ to $\mathbf{x}'$ to obtain the final transformed point $\mathbf{x}'' = \mathbf{x'} + f^{-1}(\mathbf{x}')$.
The goal is for $\mathbf{x}''$ to be close to $\mathbf{x}$, indicating that the forward and backward deformations are consistent.
We use the relative deformation error as the metric:
\begin{equation}
    \mathcal{L} = \frac{2 \|\mathbf{x}'' - \mathbf{x}\|_2} {\|\mathbf{x}' - \mathbf{x}\|_2 + \|\mathbf{x}'' - \mathbf{x}'\|_2}
\end{equation}

We compare our invertible network with the single directional deformation network of \citet{peng2022animatable}, which is a 9 hidden layer, 256 width MLP.
For the single directional MLP, we assume that the backward deformation is simply the negative of the forward deformation: $f^{-1}(\mathbf{x}) = -f(\mathbf{x})$.
We find that the average deformation error of the single directional MLP is $9.45\%$, while the deformation error of our invertible network is only $1.66\%$.
This indicates that our network is significantly better at maintaining cycle consistency. 

\section{Correspondence Visualization}
\begin{figure*}[t]
\begin{center}
   \includegraphics[width=0.65\linewidth]{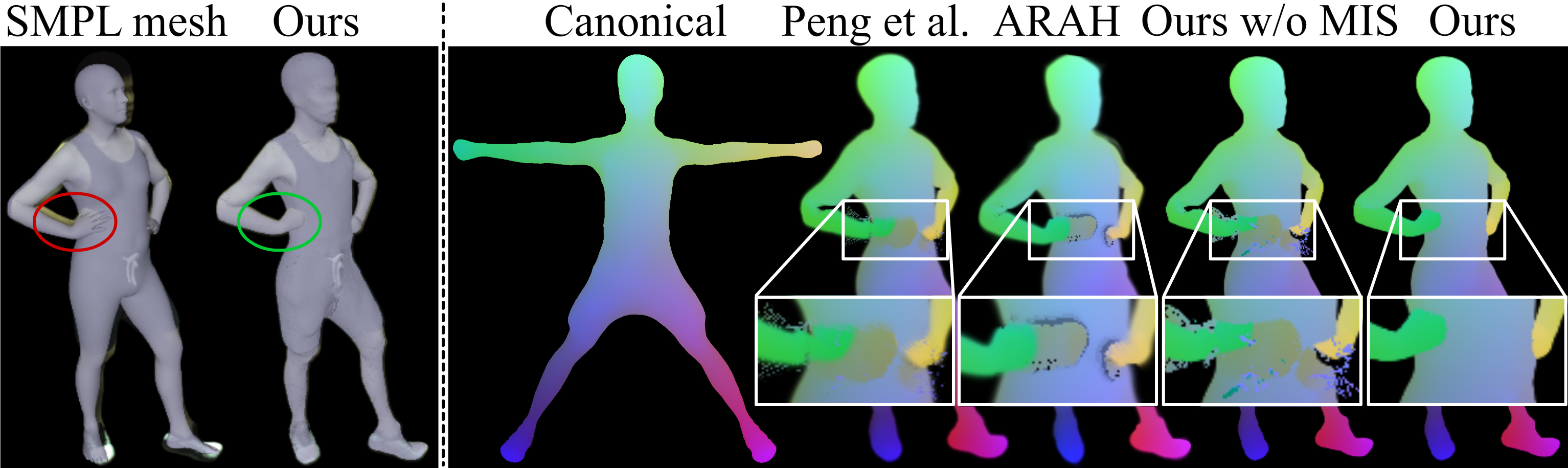}
\end{center}
\caption{Visualization of estimated correspondences.}
\label{fig:corres}
\end{figure*}

We show the estimated correspondences in Fig.\ref{fig:corres}, the results correspond to Fig.\ref{fig:geo_artifact}.
The left part shows the overlay between the mesh and the image and the right part shows the correspondences between the canonical space and the observation space, where color encodes the correspondences. 
Since the explicit mesh fits the body surface well, the computed inverse skinning weights find better correspondences between the canonical and observation spaces, and using only SMPL will lead to artifacts due to the miss matches as shown in the results of ``Ours w/o MIS".
Besides, the correspondences of other methods also mix the arms with the body. 

\begin{figure}[t]
\begin{center}
   \includegraphics[width=1.0\linewidth]{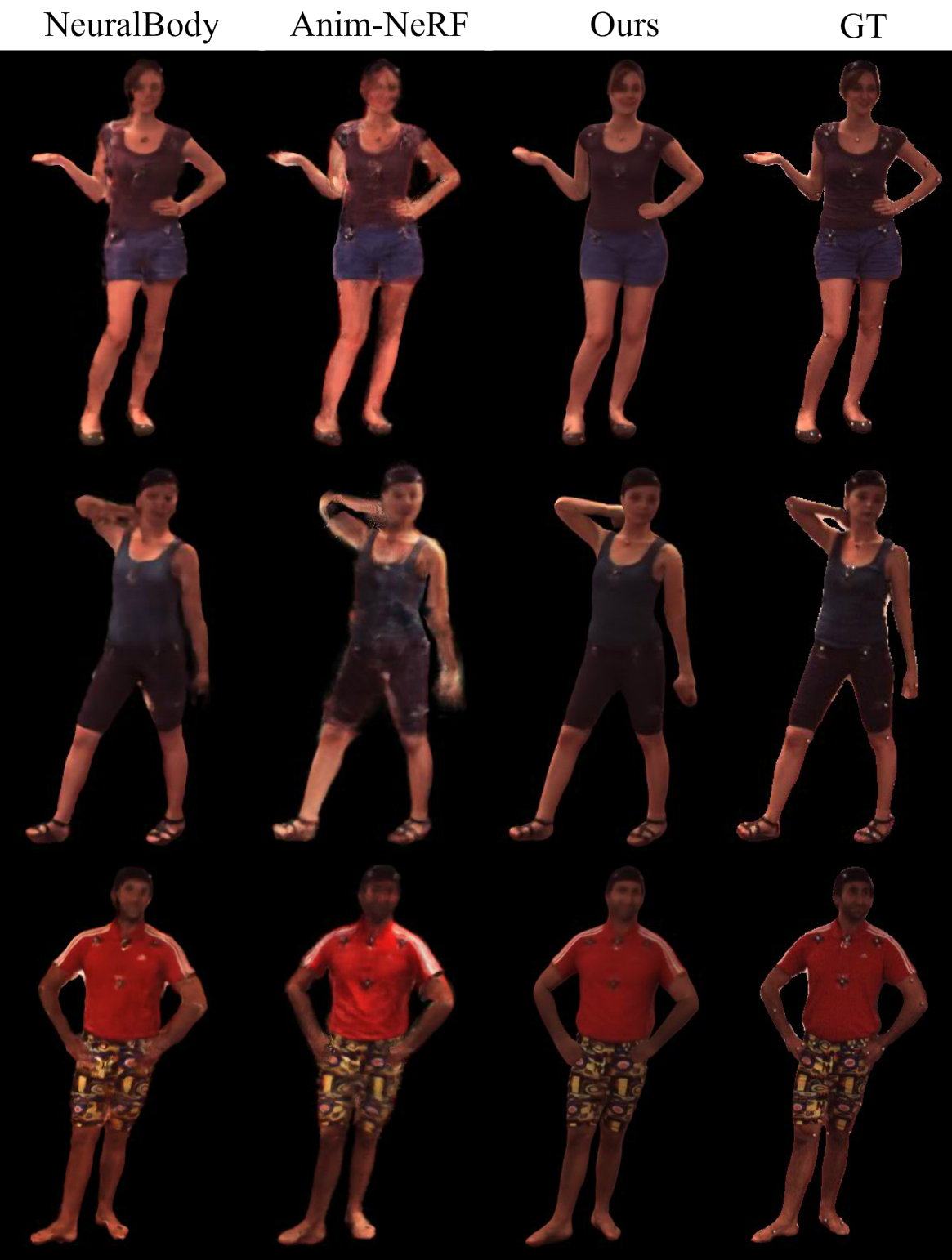}
\end{center}
\caption{Qualitative comparisons of novel pose synthesis on the Human3.6M dataset.}
\label{fig:h36m}
\end{figure}

\section{Comparisons with Non-relightable Methods}
In this section, we compare our method with state-of-the-art non-relightable methods on the Human3.6M \cite{h36m} dataset for novel view and novel pose synthesis.
We follow the test split of Anim-NeRF \cite{animatablenerf} and show quantitative results of novel view and novel pose synthesis in Tab.~\ref{tab:non-relit-sp} and Tab.~\ref{tab:non-relit-np} respectively.
The numerical results of these methods come from the recent work NPC \cite{su2023npc}.
We can see that our method achieves comparable results with state-of-the-art non-relightable methods and our results are better than NeuralBody \cite{neuralbody}, Anim-NeRF \cite{animatablenerf} and A-NeRF \cite{a-nerf}.
Note that these non-relightable methods directly predict view-dependent color, while the results of our method are rendered with the reconstructed material and lighting. 
So the results of these methods cannot be relight under novel scenes like ours.
We further show qualitative comparisons of novel pose synthesis with NeuralBody \cite{neuralbody} and Anim-NeRF \cite{animatablenerf} in Fig.~\ref{fig:h36m}.
Our method achieves better visual quality than the other two methods.

\begin{table*}[t]
\begin{center}
\scalebox{0.85}{
\begin{tabular}{cccccccccc}
\hline
\multirow{2}{*}{Method} & \multicolumn{3}{c}{S1} & \multicolumn{3}{c}{S5} & \multicolumn{3}{c}{S9} \\
                        & PSNR$\uparrow$ & SSIM$\uparrow$ & LPIPS$\downarrow$ & PSNR$\uparrow$ & SSIM$\uparrow$ & LPIPS$\downarrow$ & PSNR$\uparrow$ & SSIM$\uparrow$ & LPIPS$\downarrow$  \\
\hline
NeuralBody \cite{neuralbody} & 22.88 & 0.897 & 0.139 & 24.61 & 0.917 & 0.128 & 24.29 & 0.911 & 0.122\\
Anim-NeRF \cite{animatablenerf} & 22.74 & 0.896 & 0.151 & 23.40 & 0.895 & 0.159 & 24.86 & 0.911 & 0.145\\
A-NeRF \cite{a-nerf} & 23.93 & 0.912 & 0.118 & 24.67 & 0.919 & 0.114 & 25.58 & 0.916 & 0.126 \\
ARAH \cite{ARAH} & 24.53 & 0.921 & 0.103 & 24.67 & 0.921 & 0.115 & 25.43 & 0.924 & 0.112 \\
DANBO \cite{danbo} & 23.95 & 0.916 & 0.108 & 24.86 & 0.924 & 0.108 & 26.15 & 0.925 & 0.108 \\
TAVA \cite{tava} & \textbf{25.28} & \textbf{0.928} & 0.108 & 24.00 & 0.916 & 0.122 & 26.20 & 0.923 & 0.119 \\
NPC \cite{su2023npc} & 24.81 & 0.922 & \textbf{0.097} & \textbf{24.92} & \textbf{0.926} & \textbf{0.100} & \textbf{26.39} & \textbf{0.930} & \textbf{0.095} \\
\hline
Ours & 24.45 & 0.910 & 0.115 & 24.59 & 0.911 & 0.119 & 26.24 & 0.920 & 0.111 \\
\hline
\end{tabular}
}
\caption{Novel-view synthesis comparisons on the Human3.6M dataset.}
\label{tab:non-relit-sp}
\end{center}
\end{table*}

\begin{table*}[t]
\begin{center}
\scalebox{0.85}{
\begin{tabular}{cccccccccc}
\hline
\multirow{2}{*}{Method} & \multicolumn{3}{c}{S1} & \multicolumn{3}{c}{S5} & \multicolumn{3}{c}{S9} \\
                        & PSNR$\uparrow$ & SSIM$\uparrow$ & LPIPS$\downarrow$ & PSNR$\uparrow$ & SSIM$\uparrow$ & LPIPS$\downarrow$ & PSNR$\uparrow$ & SSIM$\uparrow$ & LPIPS$\downarrow$  \\
\hline
NeuralBody \cite{neuralbody} & 22.10 & 0.878 & 0.143 & 23.52 & 0.897 & 0.144 & 23.05 & 0.885 & 0.150 \\
Anim-NeRF \cite{animatablenerf} & 21.37 & 0.868 & 0.167 & 22.29 & 0.875 & 0.171 & 23.73 & 0.886 & 0.157 \\
A-NeRF \cite{a-nerf} & 22.67 & 0.883 & 0.159 & 22.96 & 0.888 & 0.155 & 24.16 & 0.889 & 0.164 \\
ARAH \cite{ARAH} & 23.18 & 0.903 & 0.116 & 22.91 & 0.894 & 0.133 & 24.15 & 0.896 & 0.135 \\
DANBO \cite{danbo} & 23.03 & 0.895 & 0.121 & \textbf{23.66} & 0.903 & 0.124 & 24.79 & 0.904 & 0.130 \\
TAVA \cite{tava} & \textbf{23.83} & \textbf{0.908} & 0.120 & 22.89 & 0.898 & 0.135 & 24.80 & 0.901 & 0.138 \\
NPC \cite{su2023npc} & 23.39 & 0.901 & \textbf{0.109} & 23.63 & \textbf{0.906} & \textbf{0.113} & 24.86 & \textbf{0.907} & \textbf{0.115} \\
\hline
Ours & 23.25 & 0.890 & 0.127 & 23.39 & 0.894 & 0.132 & \textbf{25.05} & 0.900 & 0.131 \\
\hline
\end{tabular}
}
\caption{Novel-pose synthesis comparisons on the Human3.6M dataset.}
\label{tab:non-relit-np}
\end{center}
\end{table*}

\begin{figure*}[t]
\begin{center}
   \includegraphics[width=1.0\linewidth]{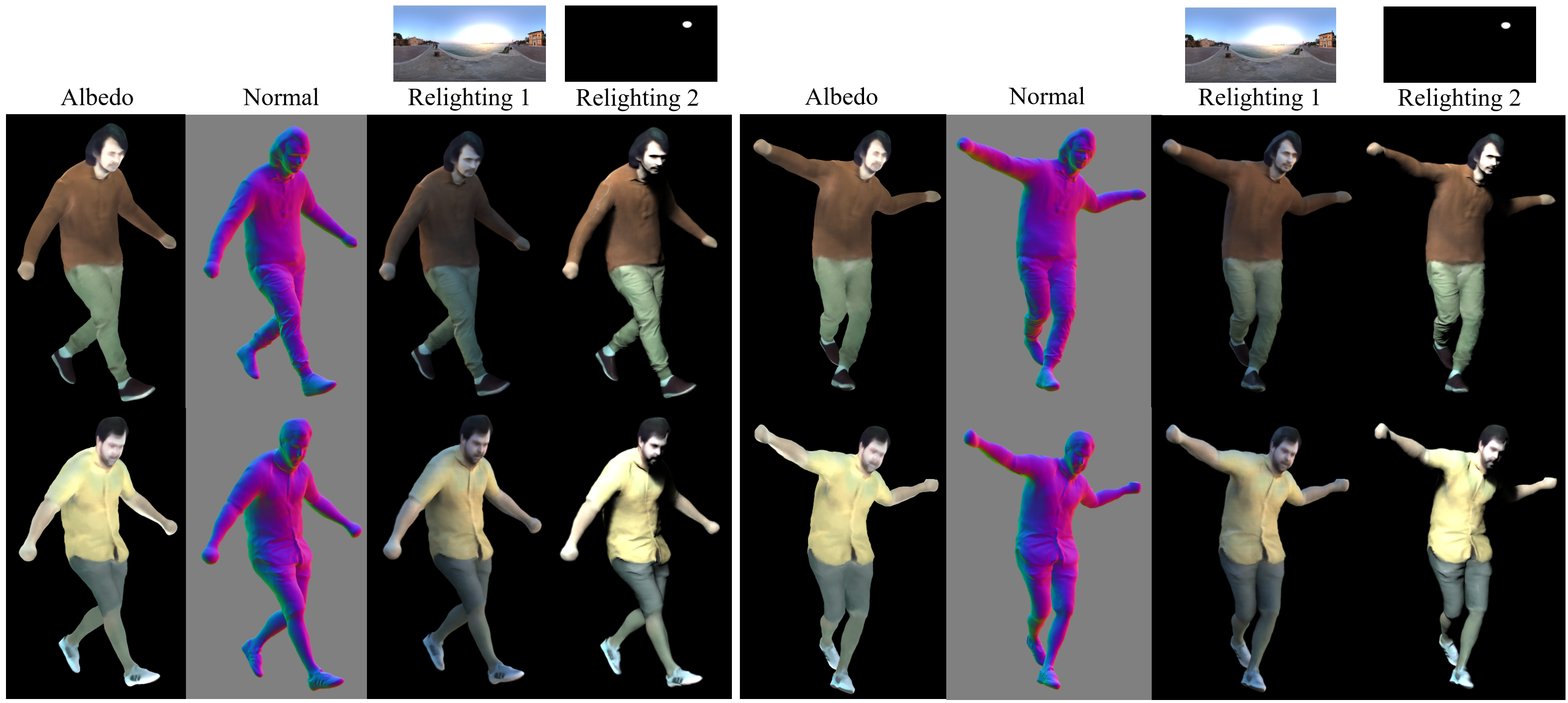}
\end{center}
\caption{Results of our technique on the DeepCap dataset. From left to right of each result: the albedo of an animated pose, the corresponding normal in this pose, and two relighting results. }
\label{fig:deepcap}
\end{figure*}

\begin{figure*}[t]
\begin{center}
   \includegraphics[width=1.0\linewidth]{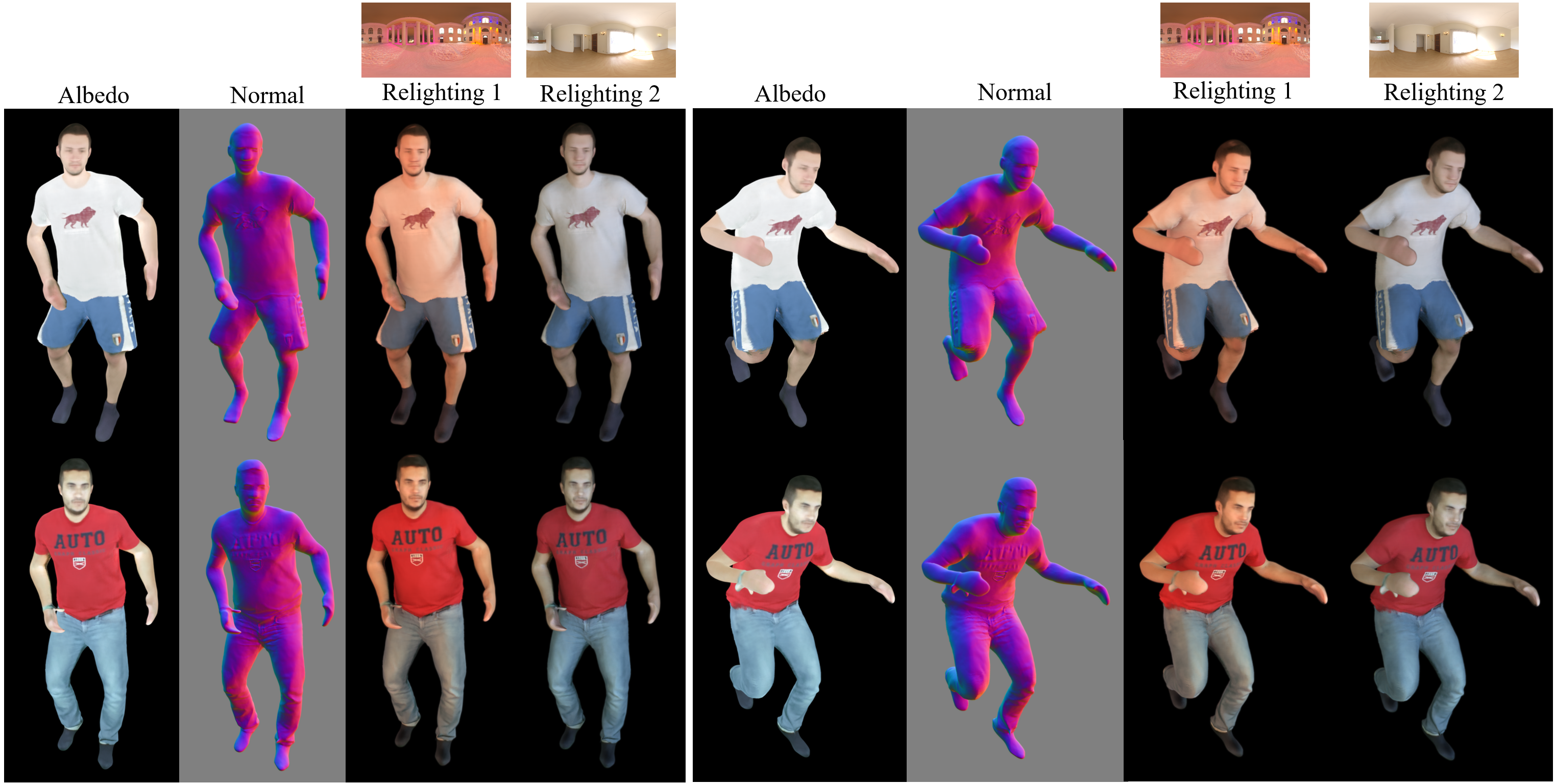}
\end{center}
\caption{Results of our technique on the PeopleSnapshot dataset. From left to right of each result: the albedo of an animated pose, the corresponding normal in this pose, and two relighting results. }
\label{fig:ps}
\end{figure*}

\begin{figure*}[t]
\begin{center}
   \includegraphics[width=0.90\linewidth]{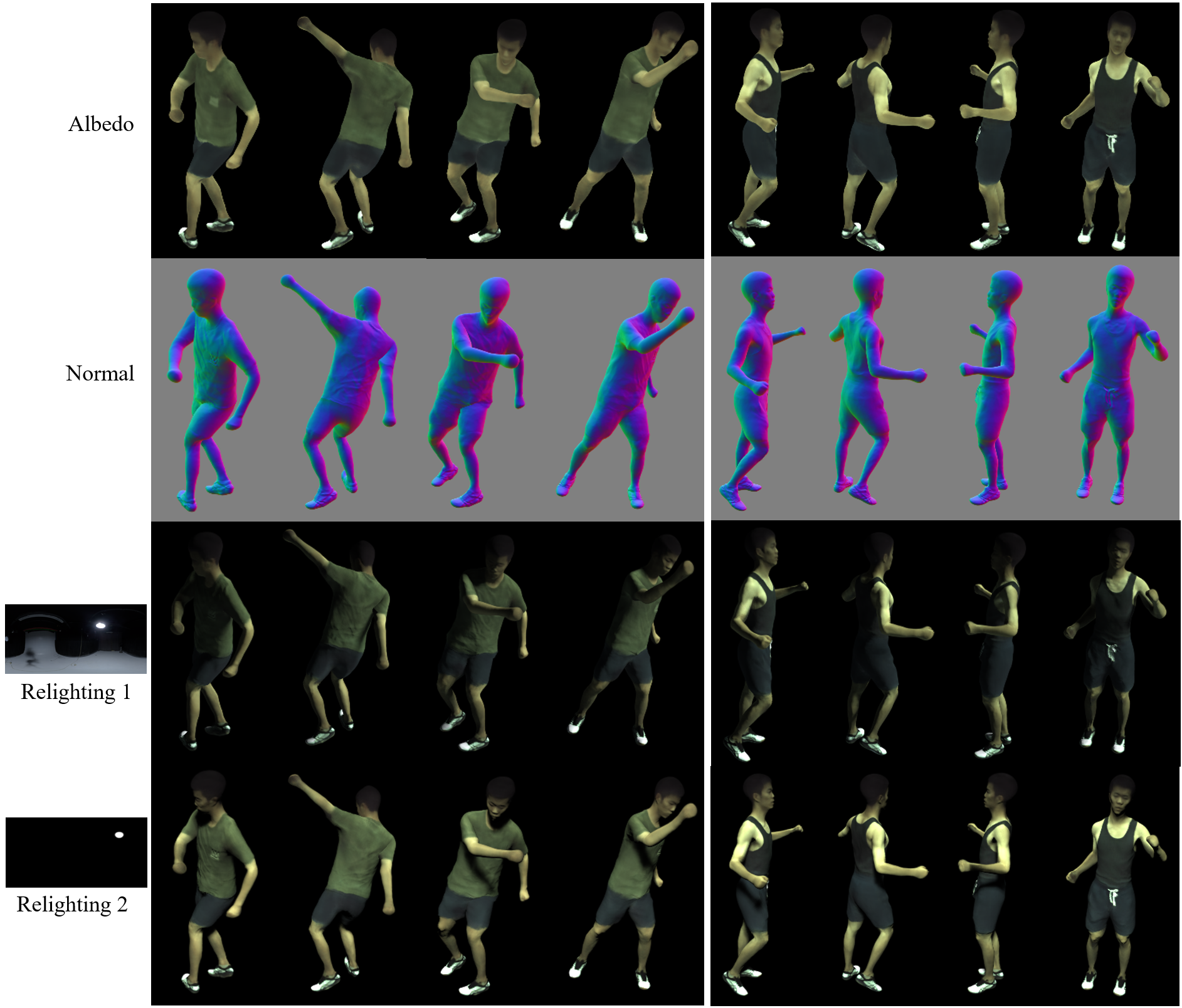}
\end{center}
\caption{Results on the ZJU-MoCap dataset with single-view input. From top to bottom: the reconstructed albedo, normal of the reconstructed geometry, and two relighting results.}
\label{fig:single_view}
\end{figure*}

\section{More Results}
In this section, we present more results of our method.
First, we show results on the DeepCap \cite{deepcap} dataset in Fig.~\ref{fig:deepcap}.
Note that our method also reconstructs geometry details like cloth wrinkles as shown in the normal maps. 
Besides, our method also works well with only single-view videos as input, we show the results of our method on the PeopleSnapshot \cite{alldieck2018video} dataset in Fig.~ \ref{fig:ps}.
Moreover, our method also produces plausible results on the ZJU-MoCap \cite{neuralbody} dataset with single input view, as shown in Fig.~\ref{fig:single_view}.
The input monocular videos from the ZJU-MoCap dataset contain 500 frames.
Note that these subjects from different datasets are captured under different lighting conditions, indicating that our method is robust to different shooting environments. 
For video results, please refer to the supplemental video.

\end{document}